\documentclass[10pt,twocolumn,letterpaper]{article}

\usepackage{cvpr}
\usepackage{times}
\usepackage{epsfig}
\usepackage{graphicx}
\usepackage{amsmath}
\usepackage{amssymb}
\usepackage{float}
\usepackage{multirow}

\usepackage[symbol]{footmisc}
\renewcommand*{\thefootnote}{\fnsymbol{footnote}}

\usepackage[breaklinks=true,bookmarks=false]{hyperref}

\cvprfinalcopy 

\def\cvprPaperID{****} 

\begin{document}

\title{
The Wisdom of MaSSeS:\\Majority, Subjectivity, and Semantic Similarity in the Evaluation of VQA
}

\author{Shailza Jolly\thanks{Shailza and Sandro share the first authorship.}\\
SAP SE, Berlin\\
TU Kaiserslautern\\
{\tt\small shailza.jolly@sap.com}
\and
Sandro Pezzelle\footnotemark[1]\\
SAP SE, Berlin\\
CIMeC - University of Trento\\
{\tt\small sandro.pezzelle@sap.com}
\and
Tassilo Klein\\
SAP SE, Berlin\\
{\tt\small tassilo.klein@sap.com}
\and
Andreas Dengel\\
DFKI, Kaiserslautern\\
CS Department, TU Kaiserslautern\\
{\tt\small andreas.dengel@dfki.de}
\and
Moin Nabi\\
SAP SE, Berlin\\
{\tt\small m.nabi@sap.com}}



\maketitle
\renewcommand*{\thefootnote}{\arabic{footnote}}
\setcounter{footnote}{0}
\begin{abstract}
We introduce \textsc{MaSSeS}, a simple evaluation metric for the task of Visual Question Answering (VQA). In its standard form, the VQA task is operationalized as follows: Given an image and an open-ended question in natural language, systems are required to provide a suitable answer. Currently, model performance is evaluated by means of a somehow simplistic metric: If the predicted answer is chosen by at least 3 human annotators out of 10, then it is 100\% correct. Though intuitively valuable, this metric has some important limitations. First, it ignores whether the predicted answer is the one selected by the Majority (\textsc{Ma}) of annotators. Second, it does not account for the quantitative Subjectivity (\textsc{S}) of the answers in the sample (and dataset). Third, information about the Semantic Similarity (\textsc{SeS}) of the responses is completely neglected. Based on such limitations, we propose a multi-component metric that accounts for all these issues. We show that our metric is effective in providing a more fine-grained evaluation both on the quantitative and qualitative level.

\end{abstract}

\section{Introduction}
Since its introduction, the task of Visual Question Answering (VQA)~\cite{antol2015vqa} has received considerable attention in the Vision and Language community. The task is straightforward: Given an image and a question in natural language, models are asked to output the correct answer. This is usually treated as a classification problem, where answers are categories that are inferred using features from image-question pairs. Traditionally, two main versions of the tasks have been proposed: One, \emph{multiple-choice}, requires models to pick up the correct answer among a limited set of options; the other, \emph{open-ended}, challenges systems to guess the correct answer from the whole vocabulary.

\begin{figure}[t!]
\begin{center}
\includegraphics[width=1\linewidth]{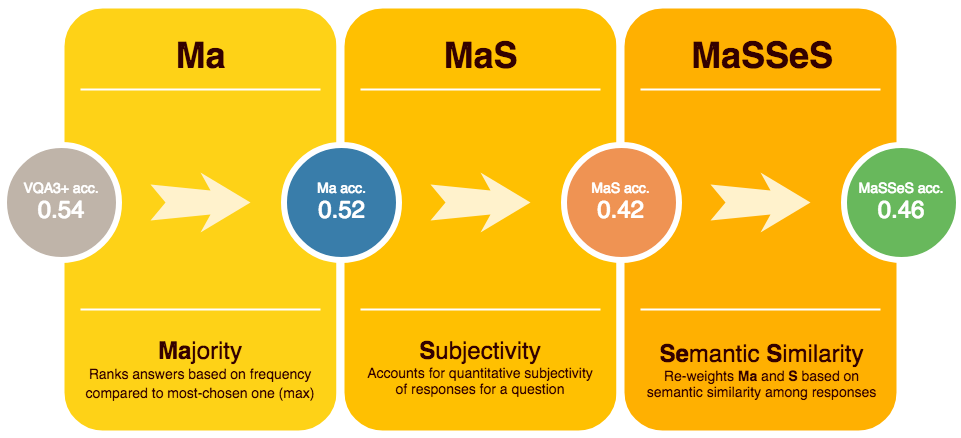}
\end{center}
\caption{Representation of \textsc{MaSSeS} and its components. In the circles, standard VQA accuracy (gray) and our \textsc{Ma} (blue), \textsc{MaS} (orange), and \textsc{MaSSeS} (green) on VQA 1.0~\cite{antol2015vqa} are reported.}\label{fig:diagram}
\end{figure}

\begin{figure*}[t!]
\begin{center}
\includegraphics[width=1\textwidth]{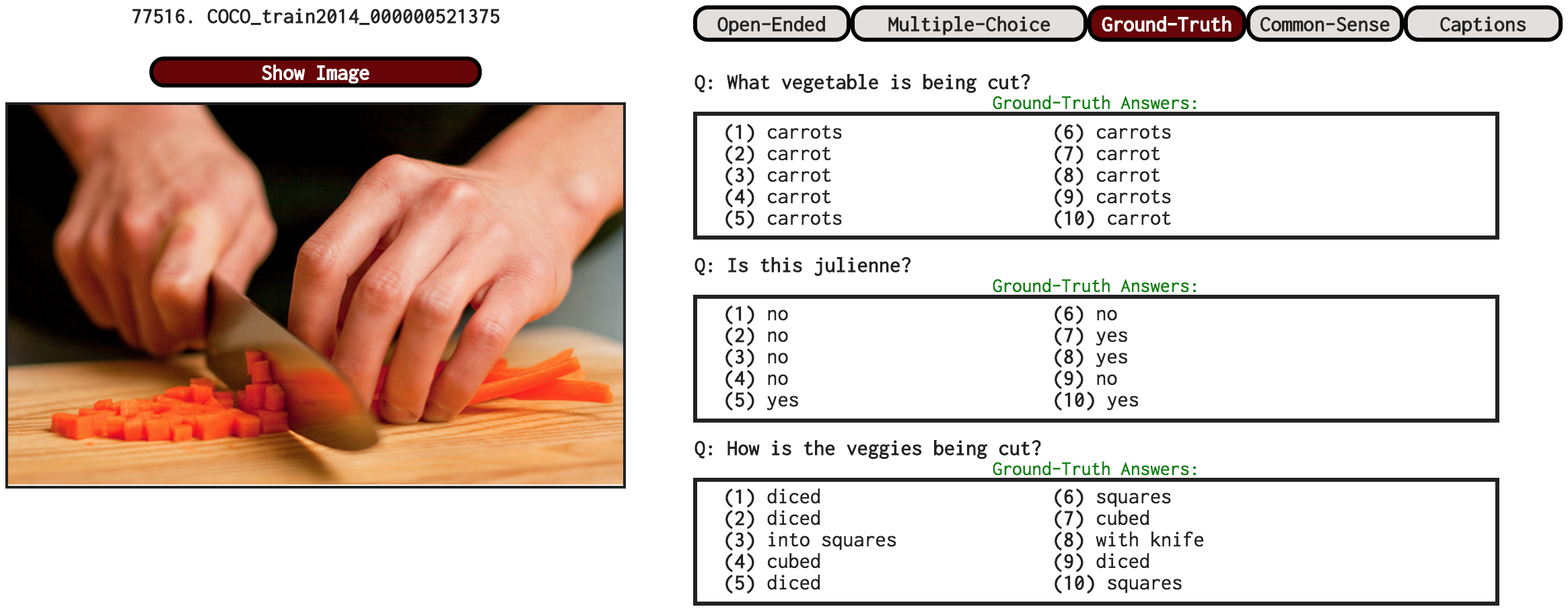}
\end{center}
\caption{Examples of VQA questions and answers in the \emph{open-ended} setting. Given the image on the left and the third question `How is the veggies being cut?', currently a model gets accuracy 100\% in case it outputs `diced' (4 occurrences), 60\% if it outputs either `cubed' or `squares' (2), 30\% for `with knife' (1), and 0\% for any other response. The overall accuracy is obtained by averaging through samples.}\label{fig:cat}
\end{figure*}

Several metrics have been proposed recently for evaluating VQA systems (see section~\ref{sec:related}), but \emph{accuracy} is still the most commonly used evaluation criterion ~\cite{antol2015vqa,fukui2016multimodal,lu2016hierarchical,xu2016ask,yang2016stacked,agrawal2017vqa,ben2017mutan,goyal2017making,yu2017multi,anderson2018bottom}. In the multiple-choice setting, where only one answer is correct, accuracy is given by the proportion of correctly-predicted cases. In the open-ended setting, accuracy is instead based on human annotations for the question:

\footnotesize
\begin{equation*}
\textsc{acc} = min(\frac{humans \ that \ said \ answer}{3},1)
\end{equation*}
\normalsize

Using the official VQA Evaluation Tool, that averages accuracy over all 10 choose 9 sets of human annotators, an answer is considered as 100\% accurate if at least 4 workers out of 10 voted for it, 90\% if the annotators were 3, 60\% if they were 2, 30\% if the answer was chosen by just one worker, 0\% in case no one opted for it.\footnote{From now on, we will report accuracy values as obtained with VQA Evaluation Tool: \url{https://github.com/GT-Vision-Lab/VQA}} Being based on the responses provided by 10 different workers, the evaluation of VQA in this setting is therefore driven by a \emph{wisdom of the crowd}~\cite{galton} criterion: The answer is `perfectly' correct if more than one third annotators agree on that, `almost' correct if the agreement involves one fifth of the workers, `a bit' correct if provided by only one worker. That is, the degree of correctness is a function of annotators agreement.

Though intuitively valuable, this metric has some important limitations. First, it ignores whether the predicted answer is the one selected by the majority of annotators or by just a smaller fraction of them. For example, in the second question in Figure~\ref{fig:cat} a model gets a 100\% accuracy by answering `yes', though this is not the most-voted option, which is `no'. Second, it does not account for the quantitative subjectivity of the responses for a given question. Based on the number of unique responses assigned by annotators, for example, the first question in Figure~\ref{fig:cat} (2 unique responses) looks intuitively less subjective compared to the third (5), but this aspect does not play any role in the evaluation. Third, information about semantic similarity of responses is completely neglected. That is, samples where the responses are very semantically similar (e.g., first question in Figure~\ref{fig:cat}) are not considered differently from cases where they are less similar (e.g., third question) or completely dissimilar (e.g., second question).

Based on such limitations, we focus on open-ended VQA and propose \textsc{MaSSeS},\footnote{Details and the code for computing \textsc{MaSSeS} will be available at the project page: \url{https://sapmlresearch.github.io/MaSSeS/}} a simple multi-component metric that jointly accounts for all these issues (see Figure~\ref{fig:diagram}). In particular, \textsc{MaSSeS} combines a Majority component (\textsc{Ma}) with a Subjectivity component (\textsc{S}) both endowed with Semantic Similarity (\textsc{SeS}). Similarly to the current evaluations, the output of the metric is a single score that measures the \emph{accuracy} in the task. By means of thorough analyses, we show that jointly considering this information is quantitatively and qualitatively better than using current evaluations. Moreover, our findings reveal that better exploiting the `wisdom of the crowd' available in human annotation is beneficial to gain a fine-grained understanding of VQA.

\section{Related Work}\label{sec:related}

In recent years, a number of VQA datasets have been proposed: VQA 1.0~\cite{antol2015vqa}, VQA-\emph{abstract}~\cite{agrawal2017vqa}, VQA 2.0~\cite{zhang2016yin,goyal2017making}, FM-IQA~\cite{gao2015you}, DAQUAR~\cite{malinowski2014multi}, COCO-QA~\cite{ren2015exploring}, Visual Madlibs~\cite{yu2015visual}, Visual Genome~\cite{krishna2017visual}, VizWiz~\cite{gurari2018vizwiz}, Visual7W~\cite{zhu2016visual7w}, TDIUC~\cite{kafle2017analysis},  CLEVR~\cite{johnson2017clevr}, SHAPES~\cite{andreas2016neural}, Visual Reasoning~\cite{suhr2017corpus}, Embodied QA~\cite{Das_2018_CVPR}. What all these resources have in common is the task for which they were designed: Given an image (either real or abstract) and a question in natural language, models are asked to correctly answer the question. Depending on the characteristics of the dataset and the models proposed, various ways to evaluate performance have been explored.

\textbf{Accuracy} is the most common metric. Traditionally, VQA is treated as a classification task, either in a multiple-choice (limited set of answers) or open-ended (whole vocabulary) setting. In the multiple-choice setting, there is just one correct (or \emph{ground-truth}) answer among a number of alternatives called \emph{decoys}~\cite{antol2015vqa,yu2015visual,zhu2016visual7w,krishna2017visual}. As such, accuracy is simply computed by counting the predictions of the model that match the ground-truth answer. What can affect the difficulty of the task in this setting is the type of decoys selected. Indeed, recent work has proposed methods to harvest more challenging alternatives on the basis of their consistency and semantic similarity with the correct response~\cite{chao2017being}. Similar approaches have been exploited in the domains of visual dialogue~\cite{das2017visual} and multiple-choice image captioning~\cite{ding2016understanding}. In the open-ended setting, accuracy can be computed in terms of \textbf{Exact Matching} between predicted and ground-truth answer~\cite{krishna2017visual,andreas2016neural,johnson2017clevr,suhr2017corpus}. Though suitable for synthetic datasets where there is just one, automatically-generated answer, this approach cannot be applied to datasets where various answers have been provided by multiple human annotators. To account for the variability among 10 crowdsourced answers,~\cite{antol2015vqa} proposed a metric which considers as 100\% correct an answer that was provided by more than 3 annotators out of 10. If 3, 2 or 1 voted for it, the model accuracy is 90\%, 60\%, and 30\%, respectively. Being simple to compute and interpret, this metric (hence, \textbf{VQA3+}) is the standard evaluation criterion for open-ended VQA~\cite{antol2015vqa,agrawal2017vqa,gurari2018vizwiz,zhang2016yin,goyal2017making}. However, it has some important limitations. (a) It ignores whether an answer that was chosen more than 3 annotators is the most frequent or not. As such, it considers it as 100\% correct even if e.g. 6 annotators converged on a different answer (see second question in Figure~\ref{fig:cat}). (b) It is heavily dependent on the number of answers for a given question. While the 3+ criterion is valid with 10 annotations, this might not be the case when, e.g., 5 or 20 answers are available. (c) It does not account for the quantitative variability among the answers. (d) There is no focus on the semantic similarity between the answers. (e) Model performance and dataset features (frequency of answers) are intertwined. That is, a perfect model cannot achieve a 100\% accuracy on the task.

\textbf{Arithmetic and Harmonic Means} are two accuracy-based metrics proposed by~\cite{kafle2017analysis}. The core idea is to compute an overall accuracy which takes into account the skewed question-type distribution observed in the TDIUC dataset. The harmonic mean-per-type accuracy (Harmonic MPT), in particular, is designed to capture the ability of a system to obtain high scores across all question-types, being skewed towards lowest performing categories. A \emph{normalized} version is also provided to better account for rare answers. Though fine-grained, these metrics are only suitable for datasets with only one ground-truth answer.

\textbf{WUPS} is a metric proposed by~\cite{malinowski2014multi} to take into account semantic similarity in the evaluation of model predictions. The core idea is that, when evaluating performance in the exact-matching setting (i.e., only one ground-truth answer), a model should not be heavily penalized if its prediction is \emph{semantically} close to the ground truth (e.g., `carton' and `box'). This intuition is implemented using Wu-Palmer similarity~\cite{wu1994verbs}, which computes the similarity between two words in terms of their longest common subsequence in the taxonomy tree. In practice, the predicted answer is considered as correct when its similarity with the ground truth exceeds a threshold, which in~\cite{malinowski2014multi} is set to either 0.9 (strict) or 0.0 (tolerant). This metric has been extended by~\cite{malinowski2015ask} to account for settings where more than one ground-truth answer is available. Two versions were proposed: In one, \textbf{WUPS-\textsc{acm}}, the overall score comes from the average of all pairwise similarities and thus considers inter-annotator agreement; in the other, \textbf{WUPS-\textsc{mcm}}, the pair with the highest similarity is taken as representative of the pattern. As observed by~\cite{kafle2017visual}, the measure of similarity embedded in WUPS has some shortcomings. In particular, it is shown to produce high scores even for answers which are semantically very different, leading to significantly higher accuracies in both~\cite{malinowski2014multi} and~\cite{ren2015exploring}. Moreover, it only works with rigid semantic concepts, making it not suitable for phrasal or sentence answers that can be found in~\cite{antol2015vqa,agrawal2017vqa,gurari2018vizwiz,zhang2016yin,goyal2017making}.

\textbf{Visual Turing Test} has been proposed as a human-based evaluation metric for VQA by~\cite{gao2015you}. Based on the characteristics of the FM-IQA dataset, whose answers are often long and complex sentences, the authors tackled the task as an answer-generation rather than a classification problem (see also~\cite{zhu2015building,wu2016value,wu2016ask,wang2017explicit,wang2017fvqa}). Given this setting, one option is to use standard metrics for the evaluation of automatically-generated language, such as BLEU~\cite{papineni2002bleu}, METEOR~\cite{lavie2007meteor}, ROUGE~\cite{lin2004rouge} or CIDEr~\cite{cider}, as~\cite{gurari2018vizwiz} did. However, these metrics turned out not to be suitable for VQA evaluation due to their inability to properly handle semantically relevant words~\cite{gao2015you}. Therefore,~\cite{gao2015you} asked humans to judge whether the generated answers were provided by a human or a model. If annotators believed the answer was `human', and thus implicitly good, the answer was considered as correct. Else, it failed the Visual Turing Test and considered as wrong. Intuitively, this evaluation procedure is very costly and heavily dependent on subjective opinions of annotators.

\begin{figure*}[t!]       
    \includegraphics[width=0.22\linewidth]{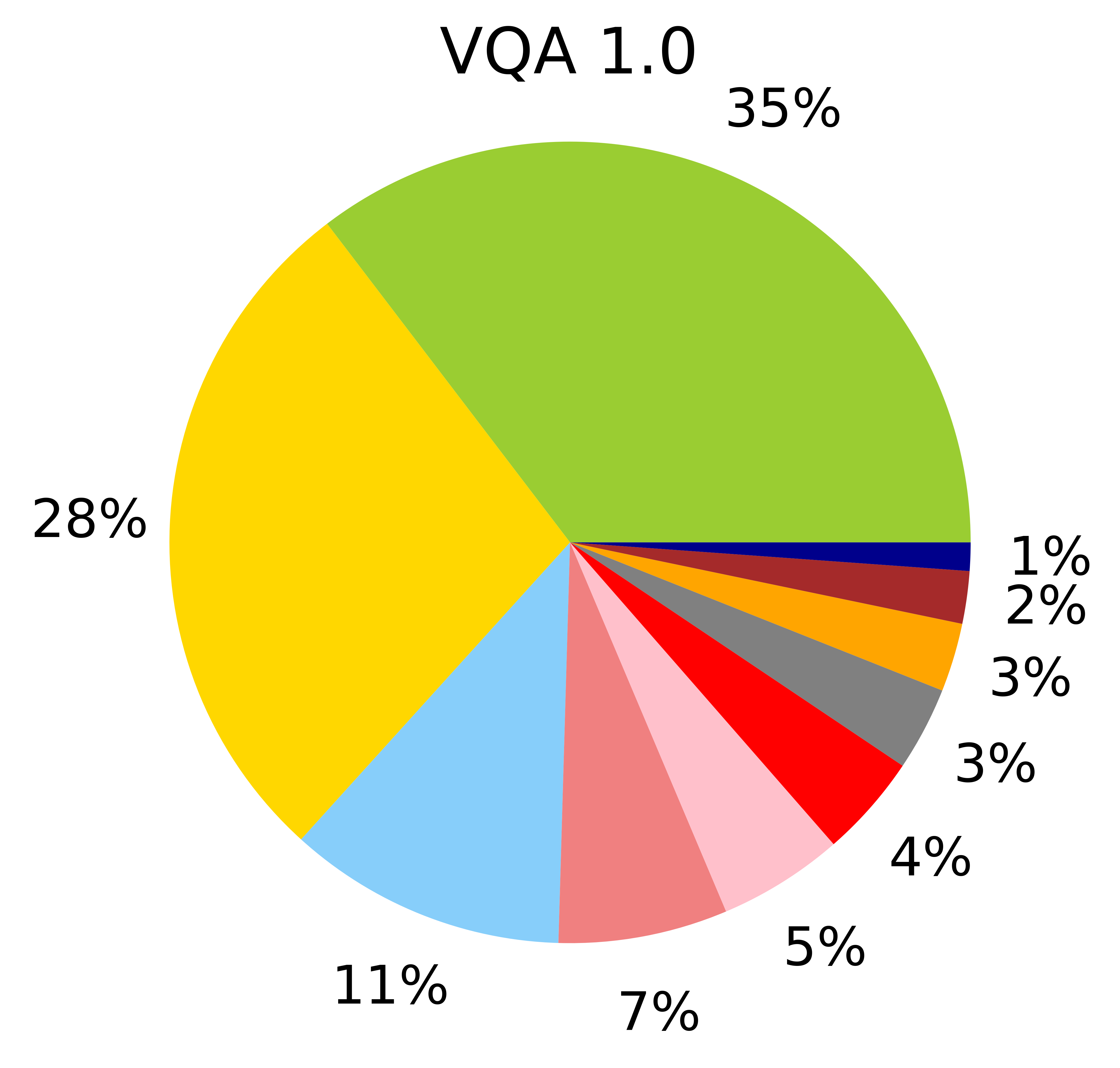}   
    \hspace{0px}
    \includegraphics[width=0.22\linewidth]{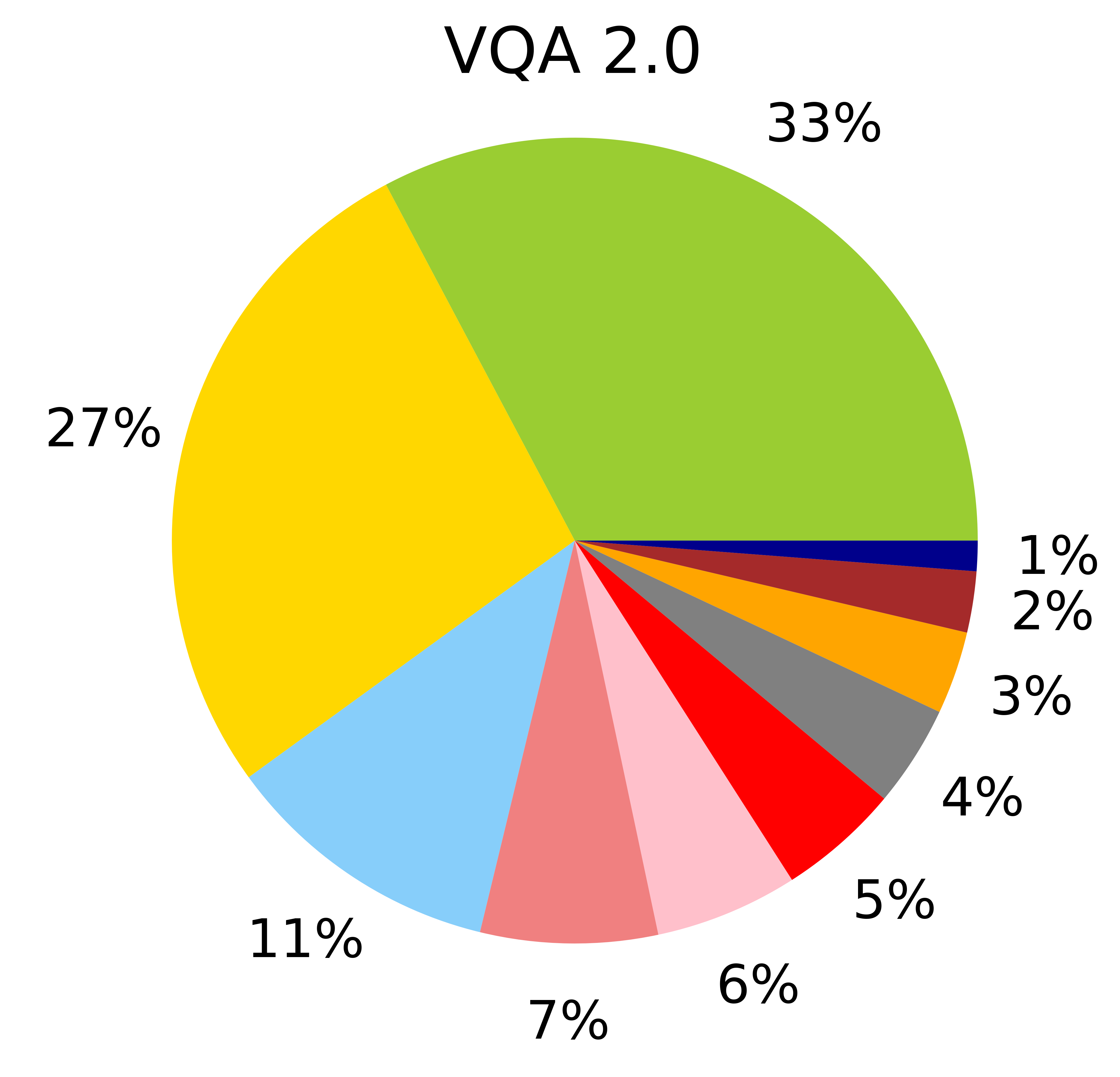}
    \hspace{0px}
    \includegraphics[width=0.22\linewidth]{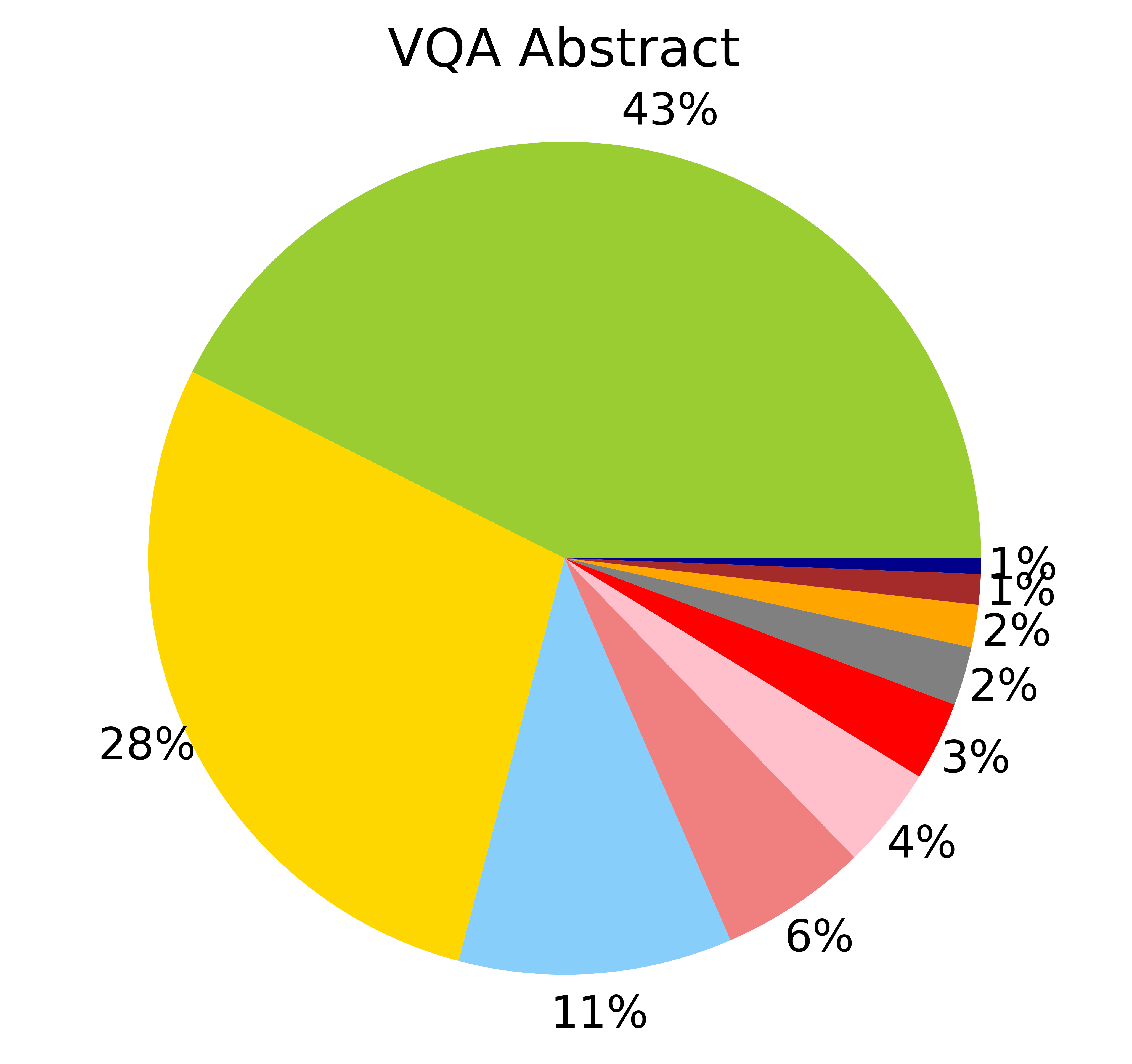}
    \hspace{0px}
    \includegraphics[width=0.22\linewidth]{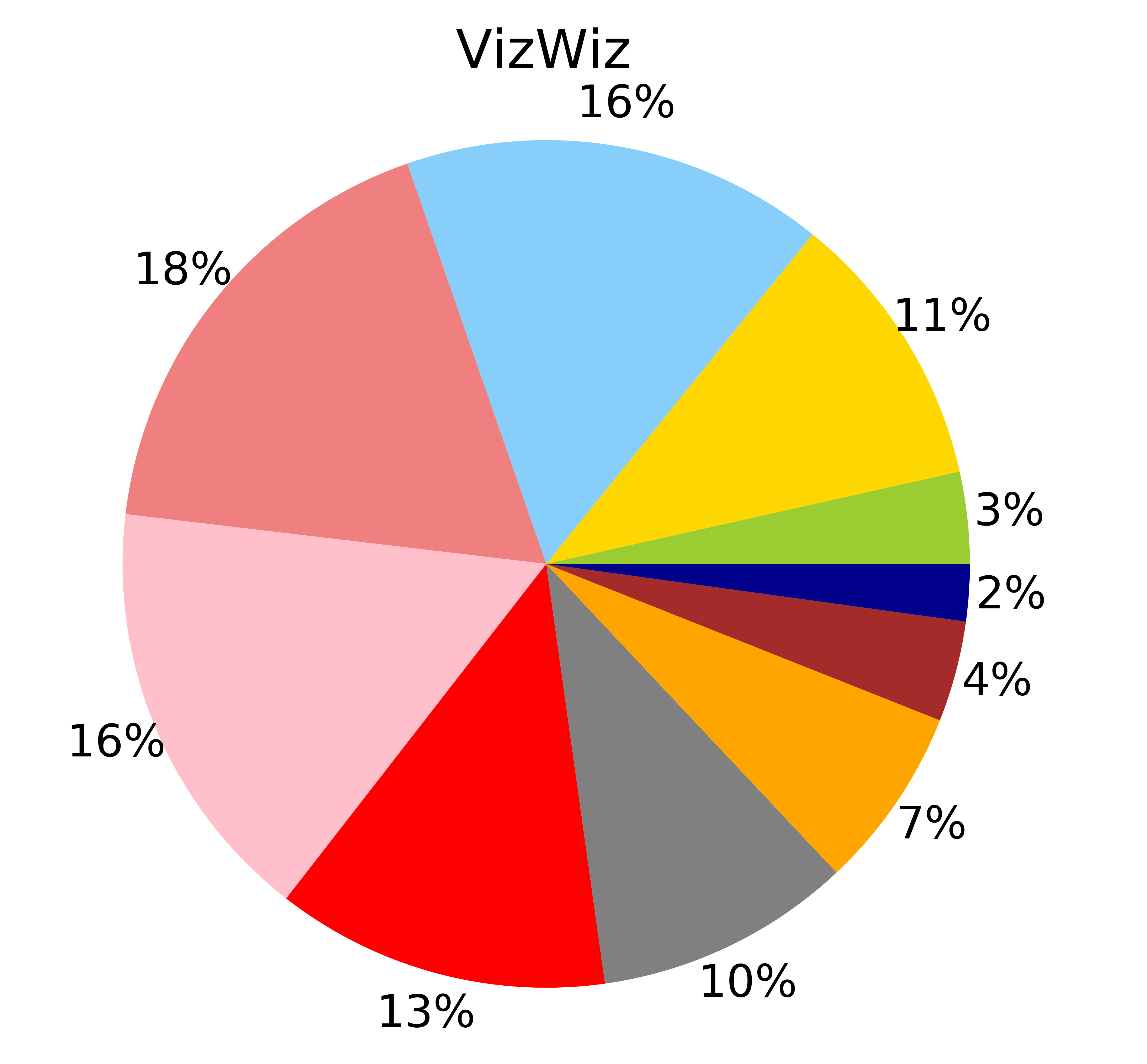}
    \hspace{0px}
    \includegraphics[width=0.06\linewidth]{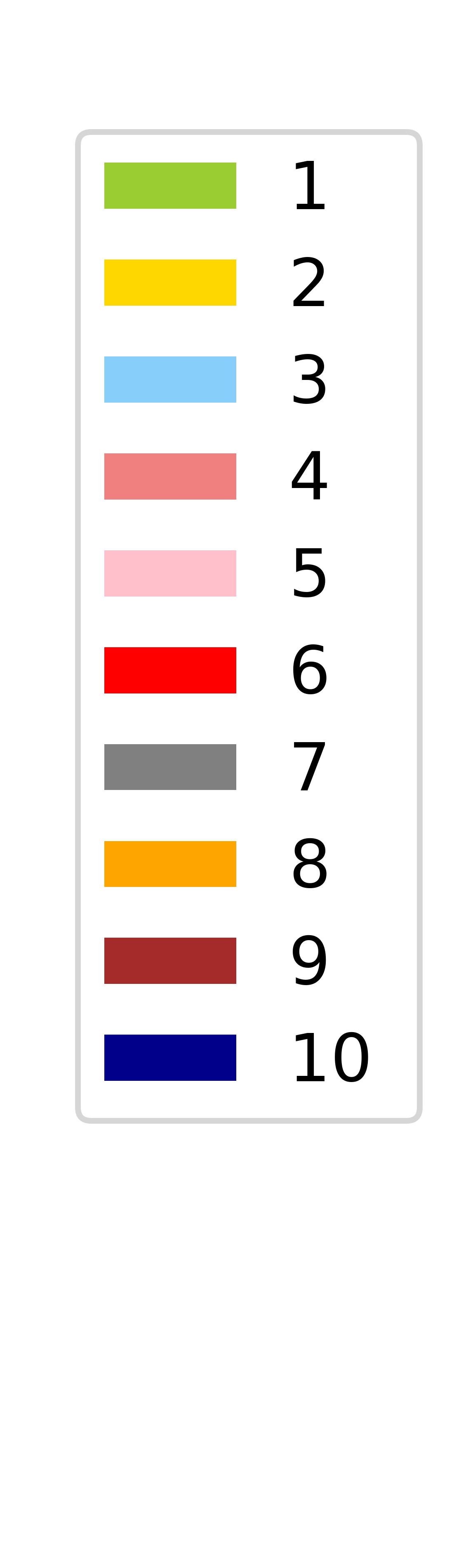}
    \caption{From left to right: Distribution of samples in the validation splits of VQA 1.0, VQA 2.0, VQA-abstract, and VizWiz with respect to number of unique answers. E.g., in 35\% of samples in VQA 1.0, all annotators converge on the same answer (3\% in VizWiz).}
    \label{fig:pies}
\end{figure*}

\textbf{Mean Rank} Finally, in the recent work by~\cite{Das_2018_CVPR} the performance of the embodied agent is evaluated via mean rank of the ground-truth answer in the predictions of the model. This implies that only one ground-truth answer is given.

\section{Our Metric}

Based on the limitations of the current metrics, we propose \textbf{\textsc{MaSSeS}}, a novel, multi-component metric for the evaluation of open-ended VQA. Each component is aimed at evaluating various aspects of either the performance of a given model or the characteristics of the dataset. In particular, one component (\textsc{Ma}) evaluates the correctness of the answer predicted by the model and is thus \emph{model-}specific. Two modules (\textsc{S}, \textsc{SeS}) evaluate the pattern of human responses for a given question and are thus \emph{data-}specific. By jointly combining these 3 modules, one single score is provided. Below, we describe and motivate each component.

\textbf{Majority (\textsc{Ma}):} It is the core component of our metric, aimed at evaluating the performance of a given model in the task. It is based on two simple assumptions: First, the most frequent answer (hence, \textsc{max}) is considered as 100\% correct regardless of its \emph{absolute} frequency. Second, all other answers receive a score which is dependent on the frequency of \textsc{max}. Given a predicted answer, the score is given by dividing its frequency by the frequency of \textsc{max}. Consider the third example in Figure~\ref{fig:cat}. If the predicted answer is `diced' (\textsc{max}), the score is 1. If it is `cubed' or `squares' (2 occurrences), the score is 0.5. If it is one among the others (1), then the score is 0.25. The method used for calculating \textsc{Ma} is reported in (1):
\begin{equation}
\textsc{Ma} = \frac{frequency \ of \ predicted \ answer}{frequency \ of \ \textsc{max}}
\end{equation}
where the numerator is an integer ranging from 0 to number of annotators (\#ann), and the denominator an integer from 1 to \#ann. \textsc{Ma} is a continuous value ranging from 0 to 1.

\textsc{Ma} overcomes some important shortcomings of the other metrics. Similarly to Exact Matching and in contrast with VQA3+, \textsc{Ma} assumes that there is \emph{always} at least one answer that is 100\% correct for the question. As a consequence, a model is allowed to achieve 100\% accuracy. Similarly to VQA3+, it modulates the score on the basis of the frequency of the answer. However, in contrast to VQA3+, our score is dependent on the frequency of \textsc{max} and not on a fixed threshold (e.g. 4). Moreover, \textsc{Ma} is \emph{continuous} (i.e., it ranges from 0 to 1) rather than discrete (VQA3+ assigns just 5 possible scores: 0\%, 30\%, 60\%, 90\%, 100\%), thus allowing a more flexible and fine-grained evaluation of the predictions.

\textbf{Subjectivity (\textsc{S}):} This component evaluates the subjectivity of a given pattern of responses on the basis of the \emph{quantitative} agreement between annotators, irrespectively of the prediction of the model. Our intuition is that highly skewed distributions would indicate more subjective and thus less reliable samples. Therefore, we should put more `trust' to distributions that reflect a high agreement compared to those where a high variability is observed. Here, we operationalize \textsc{S} in terms of Wasserstein Distance (hence, WD)~\cite{ramdas2017wasserstein}, a method applied to transportation problems using efficient algorithms like network simplex algorithm~\cite{orlin1997polynomial}. Given its ability to operate on variable-length representations, WD is more robust in comparison to other histogram-matching techniques and has been used, for example, in the domain of content-based image retrieval~\cite{rubner2000earth,rubner1998metric}. Applied to discrete probability distributions, WD (also known as Earth Mover's Distance~\cite{rubner2000earth}) is used to compute the minimum amount of \emph{work} that is needed for transforming one distribution into another. In our case, the work we measure is that required to transform a given distribution of frequencies into a uniform distribution where all elements have \textsc{max} frequency. In particular, we use WD as a measure of `reliability' of the sample, based on the observation that highly skewed distributions require a smaller amount of work (low WD) compared to `peaky' ones (high WD). This is intuitive since, in the former case, all elements are closer to the \textsc{max} than in the latter. As a consequence, patterns where all annotators converge on one single answer will get a S score equal to 1 (highest reliability), whereas uniformly-distributed patterns (i.e., all answers have frequency 1) will get 0 (no reliability at all). Consider the examples in Figure~\ref{fig:cat}. In the first and second, S is 0.55. In the third, more subjective, S is 0.33. The method used for computing \textsc{S} is shown in (2):
\begin{equation}
\textsc{S}(u,v) =  \underset{\pi\in \Gamma{u,v}}{\mathrm{inf}} \int_{R*R} |x-y| d\pi(x,y)
\end{equation}
where the formula represents the standard way for computing WD, \emph{u},\emph{v} are two different probability distributions, and $\Gamma({u,v})$ is the set of (probability) distributions. The value of \textsc{S} is further normalized to range from 0 to 1.

\begin{table*}[t!]
\begin{center}
\begin{tabular}{|l||l||l|l||l|l|l|l||l|l|l|}
\hline
\multicolumn{1}{|c||}{\textbf{dataset}} & \multicolumn{10}{c|}{\textbf{metric}}                                                                                                                                                                                       \\ \hline
                                       & \textbf{VQA3+}        & \multicolumn{2}{c||}{\textbf{WUPS}}                        & \multicolumn{7}{c|}{\textsc{\textbf{MaSSeS}}}                                                                                                       \\ \hline
\multicolumn{1}{|c||}{}                 & \multicolumn{1}{c||}{} & \multicolumn{1}{c|}{\textsc{acm}\textsubscript{0.9}} & \multicolumn{1}{c||}{\textsc{mcm}\textsubscript{0.9}} & \multicolumn{1}{c|}{\textsc{Ma}} & \multicolumn{1}{c|}{\textsc{S}} & \textsc{SeS}\textsubscript{0.7} & \multicolumn{1}{c||}{\textsc{SeS}\textsubscript{0.9}} & \multicolumn{1}{c|}{\textsc{MaS}} & \textsc{MaSSeS}\textsubscript{0.7} & \textsc{MaSSeS}\textsubscript{0.9} \\ \hline\hline
\textit{VQA 1.0}                       &  0.542                & 0.479                       & 0.642                       & 0.523                   & 0.731                  & 0.922  & 0.786                       & 0.425                    & 0.567     & 0.458    \\ \hline
\textit{VQA 2.0}                       & 0.516                & 0.441                       & 0.634                       & 0.495                   & 0.705                  & 0.907  & 0.760                       & 0.384                    & 0.545     & 0.418     \\ \hline
\textit{VQA-abstract}                  & 0.602                 & 0.532                       & 0.685                       & 0.582                   & 0.780                  & 0.944  & 0.818                       & 0.482                    & 0.618     & 0.507     \\ \hline
\textit{VizWiz}                        & 0.448                 &  0.163                           & 0.441                            & 0.444                   & 0.460                  & 0.705  & 0.541                       & 0.207                    & 0.292     & 0.227    \\ \hline
\end{tabular}
\caption{Results of VQA3+, WUPS-\textsc{acm}, WUPS--\textsc{mcm}, \textsc{MaSSeS} and its components on four VQA datasets.}\label{tab:results}
\end{center}
\end{table*}

Introducing such component allows us to take into account the subjectivity of a sample (and a dataset). This is crucial since, as shown in Figure~\ref{fig:pies}, in current datasets the proportion of samples with a perfect inter-annotator agreement (i.e., 1 unique answer) is relatively low: 35\% in VQA 1.0~\cite{antol2015vqa}, 33\% in VQA 2.0~\cite{goyal2017making}, 43\% in VQA-\emph{abstract}~\cite{agrawal2017vqa}, and only 3\% in VizWiz~\cite{gurari2018vizwiz}. Moreover, we compute this score independently from the predictions of the models, thus providing a self-standing measure for the analysis of any VQA dataset. As clearly depicted in Figure~\ref{fig:pies}, subjectivity is indeed a property of the datasets: In VizWiz, only 30\% of samples display 3 or less unique answers, whereas this percentage exceeds 70\% in the other datasets. The motivation behind proposing this component is loosely similar to~\cite{gurari2017crowdverge}, who tackle the task of predicting the degree of agreement between annotators, and very close to~\cite{yang2018visual}, who model subjectivity of samples in terms of the \emph{entropy} of the response pattern (ranging from 0 to 3.32). Compared to~\cite{yang2018visual}, we believe ours to be an essentially equivalent measure, though simpler and more intuitive. Finally, subjectivity is indirectly taken into account in WUPS-\textsc{acm}, where the score is given by the average of the pairwise distances between the elements. However, this measure mixes quantitative (frequency) and qualitative (semantic similarity) information, while S specifically focuses on the former.

\textbf{Semantic Similarity (\textsc{SeS}):} This component is aimed at evaluating the semantic similarity between the answers in the sample. The rationale is that samples where the answers are overall semantically similar should be considered as more reliable (less subjective) compared to those including semantically diverse answers. Intuitively, a pattern containing e.g. `plane', `airplane', and `aircraft' would be more consistent than one including e.g. `plane', `train', `motorbike'. We operationalize this intuition by using pre-trained word embeddings~\cite{mikolov2018advances} to re-organize the frequency distribution of the answers in the pattern. As a consequence, \textsc{SeS} can be seen as a \emph{semantics-aware} version of \textsc{S}. Technically, \textsc{SeS} is obtained as follows: (a) we compute an average representation of each answer (similarly to~\cite{chao2017being}); (b) we use these \emph{unique} representations to build a \emph{centroid} of the pattern aimed at encoding its overall semantics, irrespective of the relative frequency of the items (we want to account for the long tail of distributions); (c) we compute the cosine similarity between centroid and each unique answer; (d) we group together the answers whose cosine similarity value exceeds a given threshold, and sum their frequencies accordingly. This way, we obtain an updated frequency distribution, on the top of which S can be computed. Notably, this is the only component of \textsc{MaSSeS} that can be `adjusted'. In particular, using `strict' thresholds (e.g. 0.9) will generate lower scores compared to using more `tolerant' ones (e.g. 0.7). To illustrate, if we apply a \textsc{SeS}\textsubscript{0.9} to the examples in Figure~\ref{fig:cat}, only the reliability of the first example increases (from \textsc{S} 0.55 to \textsc{SeS} 1). However, by applying \textsc{SeS}\textsubscript{0.7}, reliability increases to 1 in all examples. Though the third question is quantitatively more subjective than the others, it becomes as reliable as them when considering its semantics. Semantic similarity is computed as in (3):
\begin{equation}
sim = cosine \ (ground \ truth \ answer, \ centroid)
\end{equation}
where for each \emph{ground truth answer, centroid} pair we obtain a similarity score \emph{sim} ranging from 0 to 1 (we set negative values to 0). Answers for which \emph{sim} is equal to or higher than a threshold \emph{t}\textsubscript{(0-1)} are grouped together by summing their frequencies. To obtain \textsc{SeS}, namely a semantics-aware measure of subjectivity, we compute (2) on the resulting distributions \emph{u}\textsubscript{sim},\emph{v}\textsubscript{sim}. To obtain the overall \textsc{MaSSeS} score, we simply compute an updated \textsc{Ma} (1) which is based on these distributions, and we further multiply it by \textsc{SeS}.

Similarly to WUPS, our metric acknowledges the importance of taking semantic similarity into account in the evaluation of VQA. However, \textsc{SeS} differs from WUPS in two main regards: (a) We use word embeddings instead of taxonomies trees, which makes our metric more flexible, intuitive, and convenient to compute. Moreover, it can account for phrasal and sentence answers. (b) As reported by~\cite{kafle2017visual}, WUPS tends to be very `forgiving' by assigning high scores to distant concepts (e.g., `raven' and `writing desk' have a WUPS score of 0.4). In contrast, word embeddings provide a more fine-grained semantic information. It is worth mentioning that, in the domain of VQA, word embeddings have been used in various ways, e.g. for selecting challenging \emph{decoys}~\cite{chao2017being}, or to implement nearest-neighbors baseline models~\cite{devlin2015exploring}. As for the procedure of aggregating various responses into one based on their semantic similarity, we were inspired by previous work on crowd consensus doing the same on the basis of various criteria~\cite{sheshadri2013square,welinder2010multidimensional}.

\begin{figure*}[t!]       
    \includegraphics[width=0.5\linewidth]{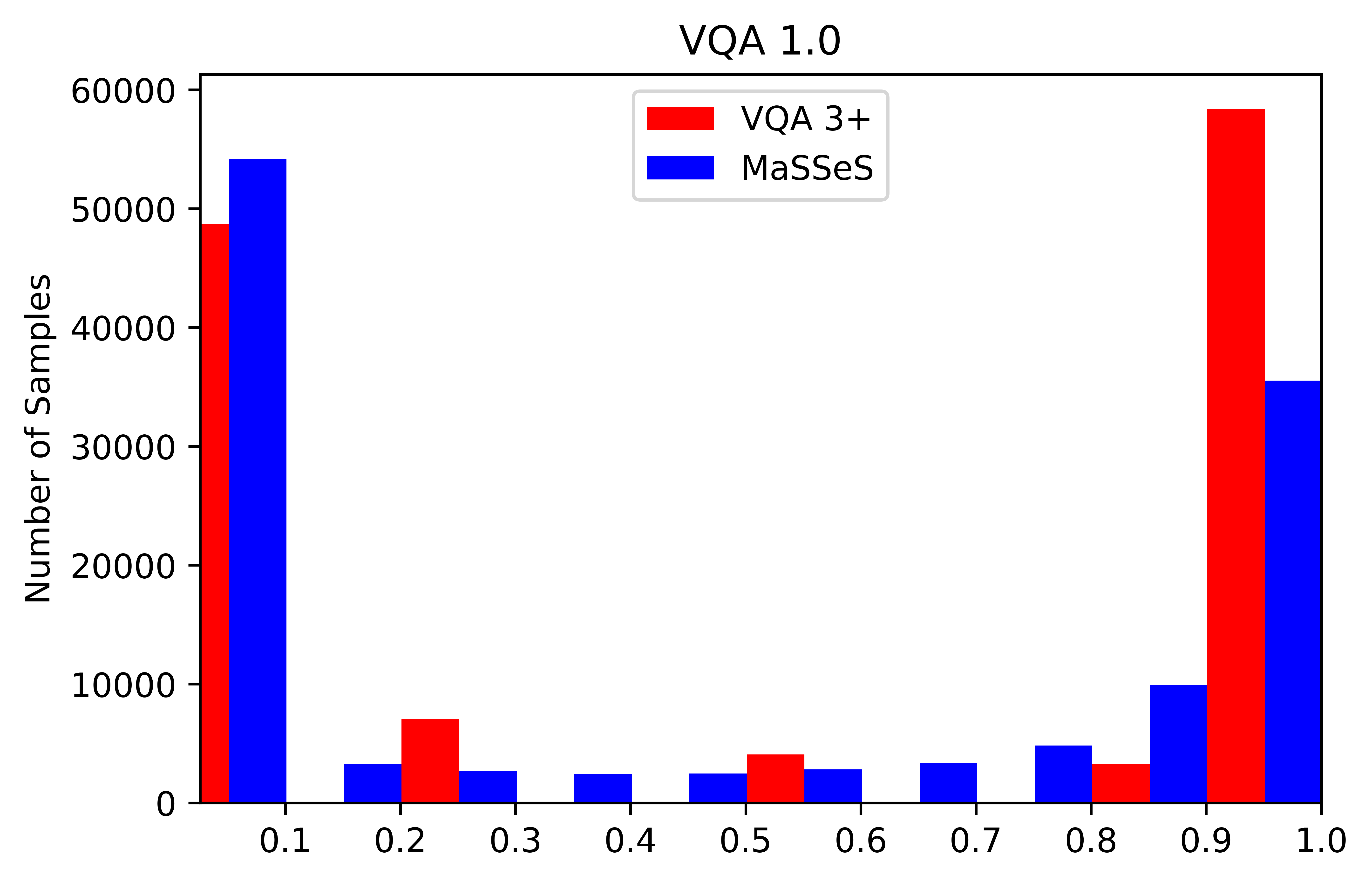}   
    \hspace{0px}
    \includegraphics[width=0.49\linewidth]{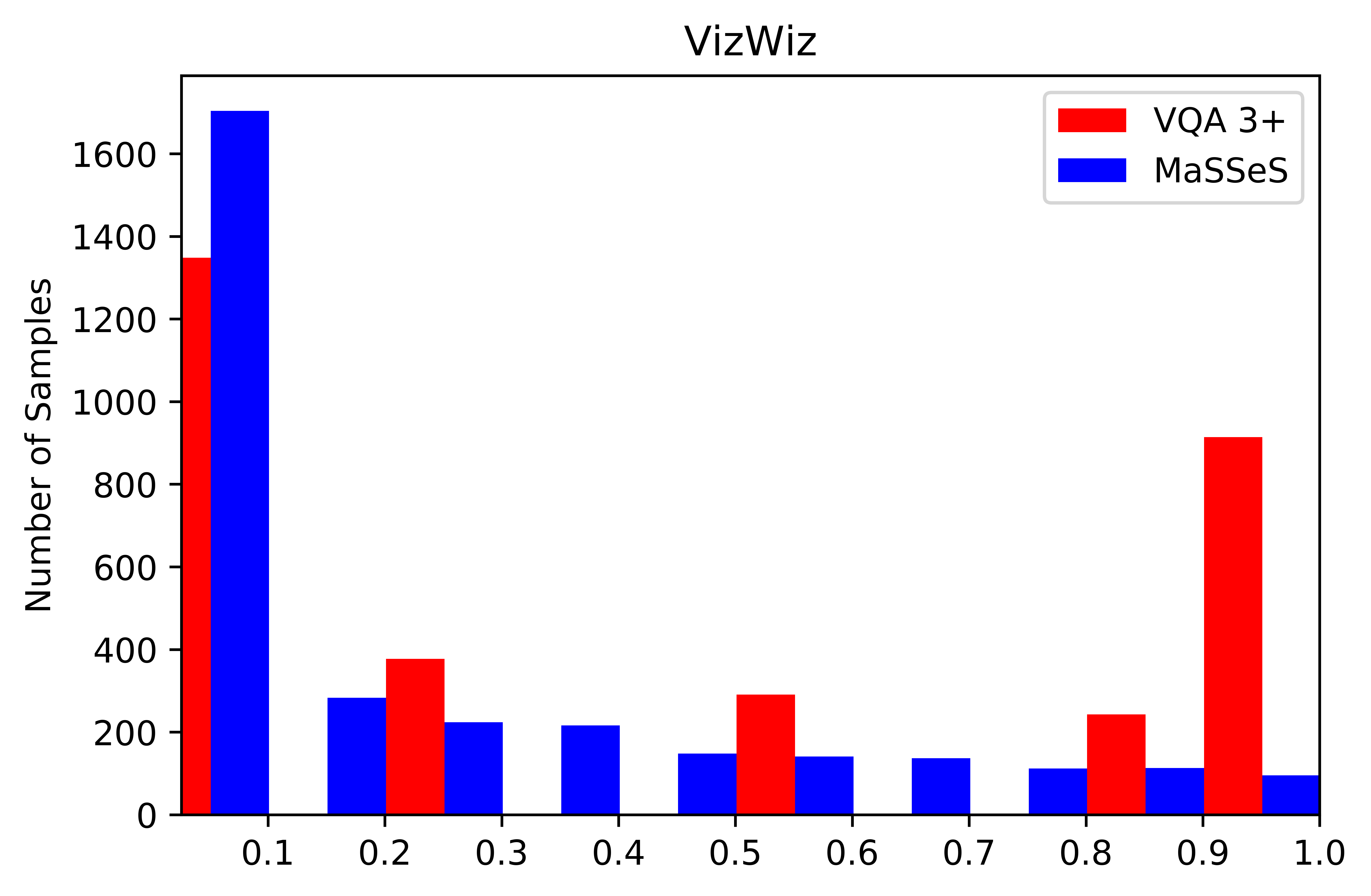}
    \caption{Comparison between VQA3+ and \textsc{MaSSeS}\textsubscript{0.9} accuracies in VQA 1.0 (left) and VizWiz (right).}\label{fig:comparison}
\end{figure*}

\begin{figure*}[t!]       
    \includegraphics[width=0.5\linewidth]{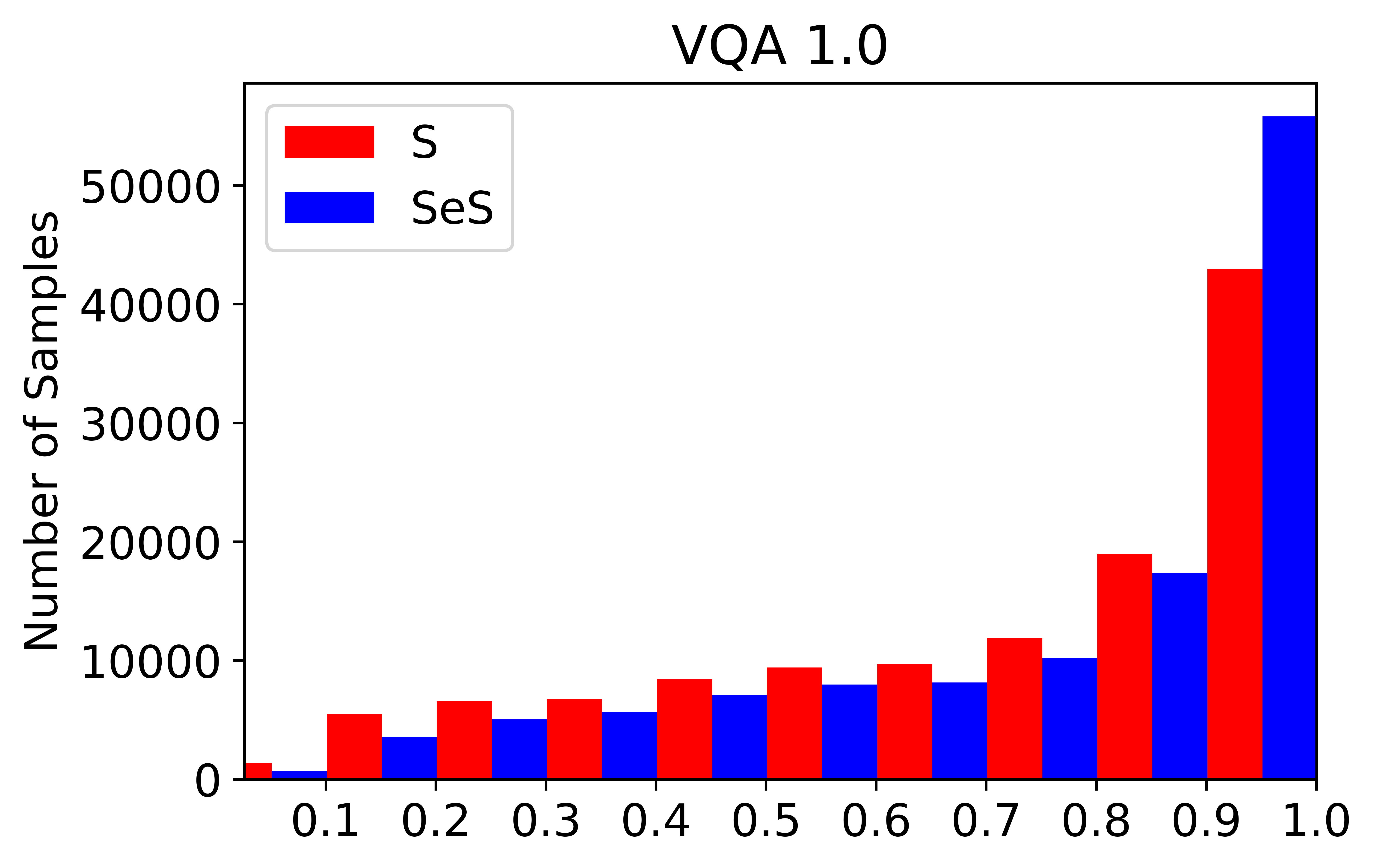}   
    \hspace{0px}
    \includegraphics[width=0.48\linewidth]{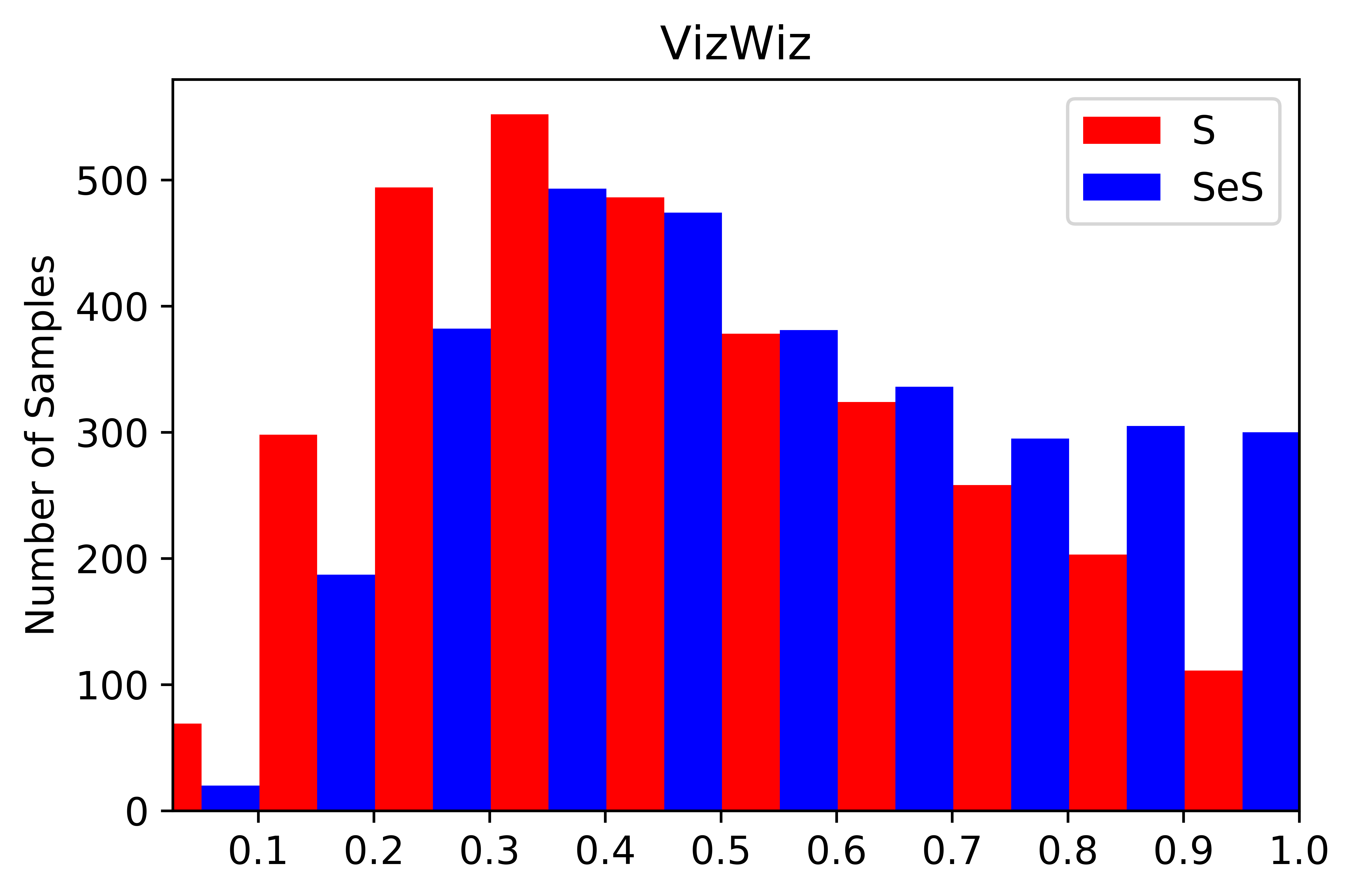}
    \caption{Distribution of Subjectivity (\textsc{S}) and Semantic Similarity (\textsc{SeS}\textsubscript{0.9}) scores in VQA 1.0 (left) and VizWiz (right).}\label{fig:sses}
\end{figure*}

\section{Experiments}

We tested the validity of our metric by experimenting with four VQA datasets: VQA 1.0~\cite{antol2015vqa}, VQA 2.0~\cite{goyal2017making}, VQA-\emph{abstract}~\cite{agrawal2017vqa}, and VizWiz~\cite{gurari2018vizwiz}. To enable a fair comparison across the datasets, for each dataset we followed the same pipeline: The standard VQA model used in~\cite{agrawal2017vqa} was trained on the training split and tested on the validation split. Model predictions were evaluated by means of three metrics: VQA3+~\cite{antol2015vqa} (using the evaluation tools), WUPS~\cite{malinowski2015ask}, and our \textsc{MaSSeS}. WUPS was tested in both its \emph{consensus} versions, i.e. \textsc{acm} and \textsc{mcm} with a threshold of 0.9. As for \textsc{MaSSeS}, we computed its overall score as well as the scores provided by each of its components. The impact of `tuning' semantic similarity is evaluated by exploring two thresholds: a strict 0.9 and a more tolerant 0.7.

\subsection{Quantitative Results}

\begin{table*}[t!]
\footnotesize
\begin{tabular}{|l|l|l|l||l|l||l|l|l|l|}
\hline
\textbf{dataset}                  & \textbf{n.} & \textbf{answers}                                                                                                                                               & \textbf{prediction}     & \textbf{\textsc{VQA}3+} & \textbf{\textsc{acm}} & \textbf{\textsc{Ma}}  & \textbf{\textsc{S}}    & \textbf{\textsc{SeS}} & \textbf{\textsc{MaSSeS}} \\ \hline\hline
\multirow{4}{*}{\emph{VQA 1.0}} & 1  & {[}yellow: 5, orange: 4, light orange: 1{]}                                                                                                           & \emph{yellow}         & 1.0   & 0.53       & 1.0 & 0.44 & 1.0    & 1.0       \\ \cline{2-10} 
                         & 2  & {[}refrigerator: 6, fridge: 4{]}                                                                                                                      & \emph{refrigerator}   & 1.0   & 0.98       & 1.0 & 0.55 & 1.0    & 1.0       \\ \cline{2-10} 
                         & 3  & \begin{tabular}[c]{@{}l@{}}{[}tennis rackets: 4, tennis racket: 2, tennis racquet: 1{]},\\ racket: 2, racquets: 1\end{tabular}                        & \emph{tennis rackets} & 1.0   & 0.98       & 1.0 & 0.33 & 0.67   & 0.67      \\ \cline{2-10} 
                         & 4  & \begin{tabular}[c]{@{}l@{}}{[}hot dogs: 5, hot dog: 2, hot dogs and fries: 1,\\ hot dog fries: 1, hot dog and onion rings: 1{]}\end{tabular}          & \emph{hot dog}        & 0.60  & 0.70       & 0.4 & 0.44 & 1.0    & 1.0       \\ \hline\hline
\multirow{4}{*}{\emph{VizWiz}}  & 1  & \begin{tabular}[c]{@{}l@{}}{[}christmas tree: 6, tree: 1, chritmas tree shaped\\ santaclauses: 1, christmas tree santas: 1{]}, santas: 1\end{tabular} & \emph{christmas tree} & 1.0   & 0.70       & 1.0 & 0.55 & 0.89   & 0.89      \\ \cline{2-10} 


& 2  & \begin{tabular}[c]{@{}l@{}}white: 6, {[}green: 2, light green: 1,\\ very light green: 1{]}\end{tabular}                                               & \emph{white}          & 1.0   & 0.62       & 1.0 & 0.55 & 0.55   & 0.55      \\ \cline{2-10} 

& 3  & \begin{tabular}[c]{@{}l@{}}{[}ginger peach: 5, ginger peach tea: 2,\\ ginger peach herbal tea: 1{]}, unanswerable: 2\end{tabular}                     & \emph{unanswerable}   & 0.60  & 0.20       & 0.4 & 0.44 & 0.77   & 0.19      \\ \cline{2-10} 
                         & 4  & \begin{tabular}[c]{@{}l@{}}{[}beef: 5, beef flavored broth: 2, beef flavored: 1,\\ beef flavor: 1, this beef flavor: 1{]}\end{tabular}                & \emph{unanswerable}   & 0.0   & 0.0       & 0.0 & 0.44 & 1.0    & 0.0       \\ \hline
\end{tabular}
\caption{Examples from the validation splits of VQA 1.0 (top) and VizWiz (bottom). For each example, we report the pattern of answers provided by annotators (unique answer: frequency), the prediction of the model, and the scores (note that \textsc{acm}, \textsc{SeS}, \textsc{MaSSeS} are computed using threshold 0.9). Answers that are grouped together by \textsc{SeS} are included in square brackets.}\label{tab:qual}
\end{table*}

Results are reported in Table~\ref{tab:results}. Note that columns VQA3+, WUPS-\textsc{acm}, WUPS-\textsc{mcm}, \textsc{Ma}, \textsc{MaS}, and \textsc{MaSSeS} are \emph{accuracies}, while \textsc{S} and \textsc{SeS} are \emph{reliability scores}. As can be noted, accuracies obtained with both versions of \textsc{MaSSeS} are generally lower compared to those of VQA3+, with the drop being particularly accentuated for VizWiz. This can be observed in Figure~\ref{fig:comparison}, which compares the distributions of accuracies scored by VQA3+ and \textsc{MaSSeS}\textsubscript{0.9} in VQA 1.0 (left) and VizWiz (right). As can be seen, the scores produced by our metric (blue) are `distributed' across the x-axis (from 0 to 1), while those produced by VQA3+ (red) are grouped into 5 `classes'. Moreover, we observe that our metric is much more reluctant to output score 1. Part of this differences can be explained by looking at the values of \textsc{Ma} (Table~\ref{tab:results}), which are slightly lower than those of VQA3+ due to their finer-grained nature (recall that if an element is not \textsc{max} it is not considered as 100\% correct by \textsc{Ma}). This drop is further accentuated by multiplying \textsc{Ma} by either \textsc{S} (to obtain \textsc{MaS}) or \textsc{SeS} (to obtain \textsc{MaSSeS}). Since the values of these components cannot exceed 1, the resulting score will be lowered according to the degree of subjectivity of the dataset.

Bearing this in mind, it is worth focusing on the scores of \textsc{S} and \textsc{SeS} in each dataset. As reported in Table~\ref{tab:results}, \textsc{S} is relatively high for the first three datasets (ranging from 0.70 to 0.78), extremely low for VizWiz (0.46). These numbers, in line with the descriptive statistics depicted in Figure~\ref{fig:pies}, clearly indicate that answers in VizWiz are extremely skewed, with annotators rarely agreeing on the same answer(s). This information can also be observed in Figure~\ref{fig:sses}, which depicts the distribution of \textsc{S} (red bars) and \textsc{SeS}\textsubscript{0.9} (blue bars) in VQA 1.0\footnote{We plot VQA 1.0 as representative of the three VQA-based datasets, which display very similar patterns.} (left) and VizWiz (right). As can be noticed, \textsc{S} in VQA is relatively high, with most of the answers being grouped in the rightmost bars (0.8 or more). In contrast, we observe an almost normal distribution of \textsc{S} in VizWiz, with very few answers being scored with high values. When injecting semantic information into subjectivity (\textsc{SeS}\textsubscript{0.9}), however, the distribution changes. Indeed, we observe much less cases scored with extremely low values and much many cases with high values. In numbers, this is reflected in an overall increase of 8 points from \textsc{S} (0.46) to \textsc{SeS} (0.54). A similar pattern is also observed in VQA 1.0 (+5 points). It is worth mentioning that using a lowest similarity threshold (0.7) makes the increase between \textsc{S} and \textsc{SeS} even bigger. This, in turn, makes the \textsc{MaSSeS} score significantly higher and comparable to VQA3+ in the three VQA-based datasets (not for VizWiz).

As for WUPS, we observe that \textsc{acm} scores are significantly lower than VQA3+ ones, while \textsc{mcm} ones are generally higher. This is intuitive since \textsc{mcm} only considers the most similar answers, while \textsc{acm}, similarly to ours, considers the whole set. Compared to our metric, we notice that \textsc{acm}\textsubscript{0.9} scores are somehow in between those of \textsc{MaSSeS}\textsubscript{0.7} and \textsc{MaSSeS}\textsubscript{0.9} in the VQA-based datasets. In contrast, they are very different in VizWiz, where our metric versions `outperform' \textsc{acm}\textsubscript{0.9} by around 13 and 7 points, respectively. We believe this gap is due to the main differences between WUPS and \textsc{MaSSeS}: (a) In WUPS the predictions of the model are intertwined with the properties of the data, while in ours the two components are disentangled. (b) The type of semantic similarity used by \textsc{MaSSeS} and its role in the metric allows capturing finer-grained relations between the answers compared to taxonomy trees.

\subsection{Qualitative Results}

To better understand the functioning of our metric, we analyze several cases extracted from the validation splits of VQA 1.0 and VizWiz (see Table~\ref{tab:qual}). Starting from VQA 1.0, we notice that examples 1 and 2 are considered as 100\% correct by both VQA3+ and \textsc{MaSSeS}. The former metric assigns this score because `yellow' and `refrigerator' have frequency equal to or greater than 4. As for \textsc{MaSSeS}, this score is produced because (a) the two answers have \textsc{max} frequency, and (b) the \textsc{SeS} score assigned to the response pattern is the highest (i.e. 1.0) due to their semantic consistency. That is, all the answers are grouped together since their cosine similarity with the centroid is equal or greater than 0.9. Notably, \textsc{acm} produces a similar score in example 2, but very different (i.e., much lower) in example 1, though the words involved are semantically very similar (very similar colors). Moving to example 3, we observe that \textsc{MaSSeS} assigns a lower score (0.67) compared to VQA3+ (1.0) since \textsc{SeS} makes a fine-grained distinction between generic  `rackets' and specific ones (i.e., for `tennis'). This proves the validity and precision our semantic similarity component, especially in comparison with \textsc{acm}, whose high score does not account for such distinction (0.98). As for example 4, the score output by \textsc{MaSSeS} (1.0) turns out to be higher than both VQA3+ (0.6) and \textsc{acm} (0.7) due to the extremely high semantic consistency of the answers.

\begin{figure*}[t!]
\centering
   \includegraphics[width=0.65\linewidth]{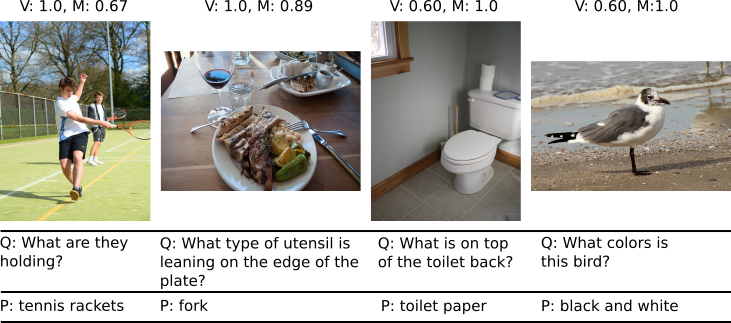}
    \caption{Left: Two examples where VQA3+ (V) outputs higher scores than \textsc{MaSSeS} (M). Right: Two examples with the opposite pattern.}
    \label{fig:vs}
\end{figure*}

\begin{figure}[b!]
\centering
    \includegraphics[width=0.49\linewidth]{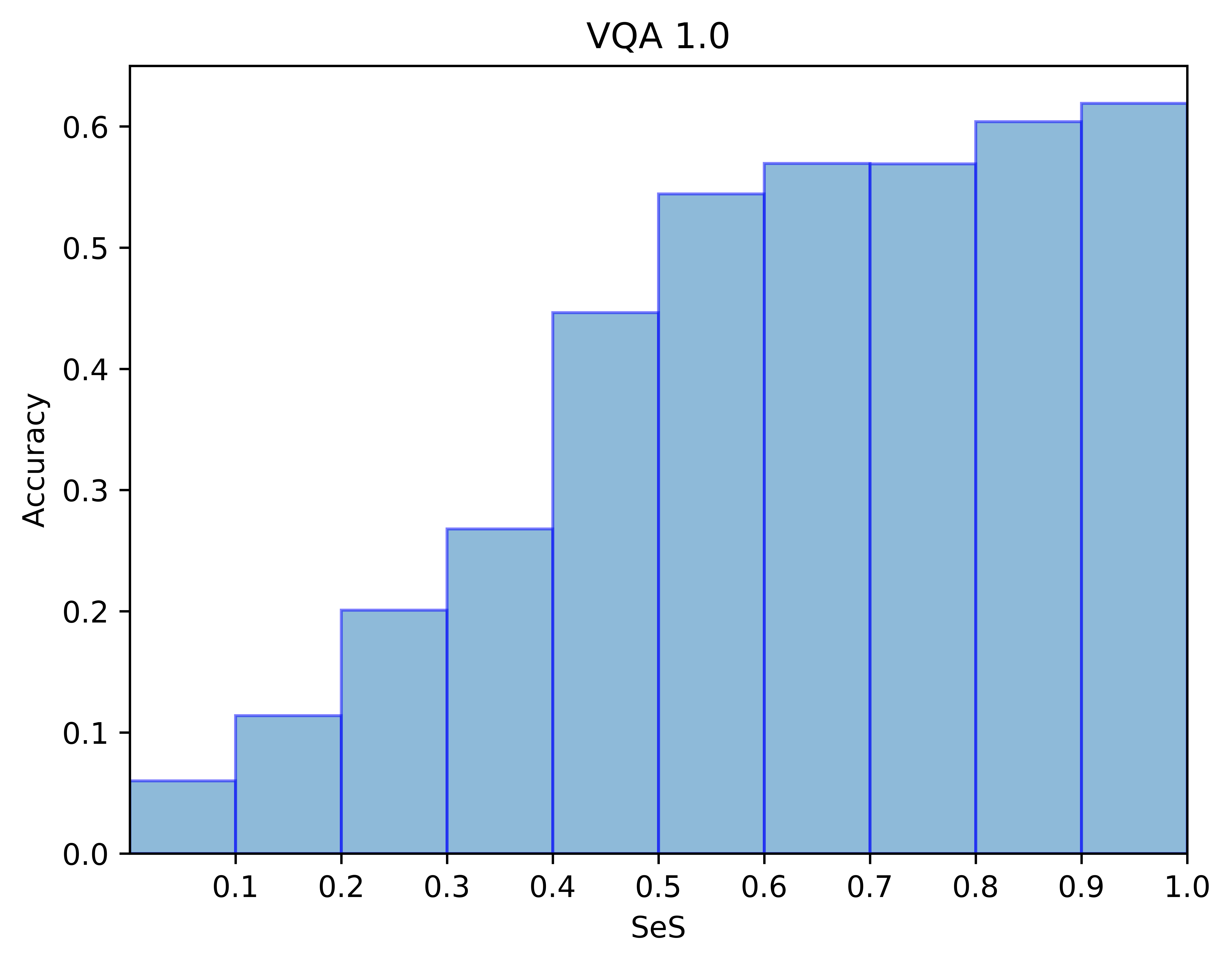}   
    \hspace{0px}
   \includegraphics[width=0.49\linewidth]{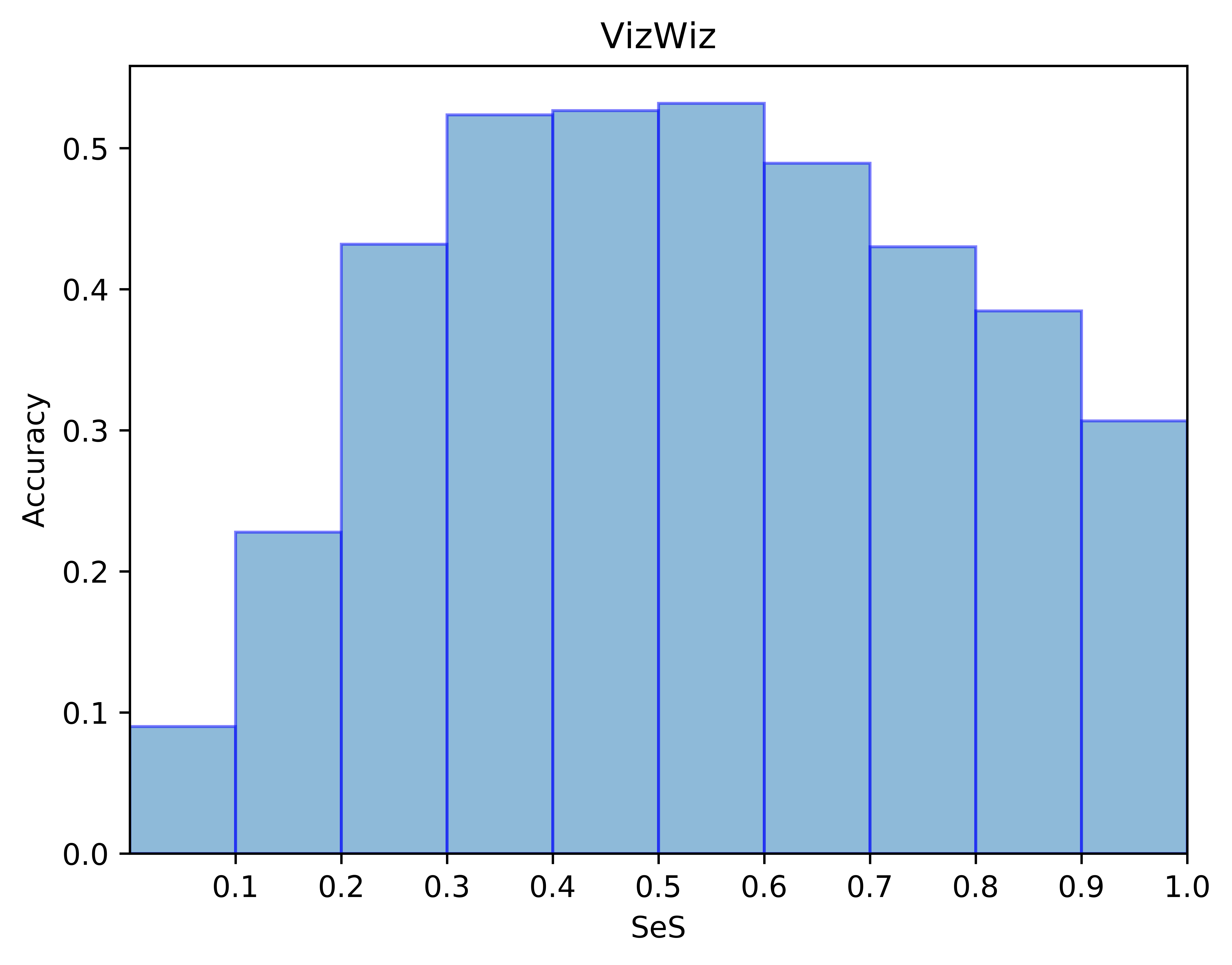}
    \caption{Distribution of accuracy produced by VQA3+ against \textsc{SeS} values in VQA 1.0 (left) and VizWiz (right).}
    \label{fig:diffbars}
\end{figure}

As for VizWiz, we observe that examples 1 and 2, which receive highest accuracy from VQA3+, are assigned a lower score by \textsc{MaSSeS}. In the former case, the drop is minor due to the high reliability of the pattern; in the latter, the drop is bigger since the predicted answer, `white', appears in a pattern where the other responses are semantically very similar to each other and thus grouped together by \textsc{SeS}. That is, the items in the long tail of the distribution, though not \emph{quantitatively} dominant, are \emph{semantically} prevalent in the pattern. As such, the reliability of the pattern is only partial, and lowers the overall score. As for example 3, VQA3+ assigns a relatively high score to the prediction (0.60), while \textsc{MaSSeS} (as \textsc{acm}) penalizes this choice mainly due to the non-\textsc{max} nature of the predicted answer, though the pattern has a high reliability due the semantic consistency of the alternatives (all grouped together by \textsc{SeS}). Finally, in example 4 the prediction of the model (`unanswerable') is not present in the pattern and thus scored 0 by all metrics. However, it is worth mentioning that, according to \textsc{SeS}, this pattern is highly reliable due to the high semantic consistency of its elements. As a consequence, a model predicting e.g. `beef' would get 1.0 by \textsc{MaSSeS}, but only 0.5 by \textsc{acm}.

To further understand the qualitative difference between VQA3+ and \textsc{MaSSeS}, we analyze several cases from VQA 1.0 (see Figure~\ref{fig:vs}) where the former metric outputs a higher score than the latter (left), and \emph{vice versa} (right). In the two leftmost examples, the higher values produced by VQA3+ seem intuitively more correct than those output by \textsc{MaSSeS}, whose scores are affected by a valuable but somehow strict semantic criterion which penalizes the presence of other answers in the pattern. In contrast, the higher accuracies produced by \textsc{MaSSeS} in the rightmost cases look intuitively better than those by VQA3+. In these cases, the subjectivity of the pattern is compensated by the high semantic consistency among the answers, which makes \textsc{MaSSeS} to output the highest score. Overall, it is straightforward that taking semantics into account allows our metric to produce finer-grained evaluations.

\section{Evaluating Dataset `Feasibility' with \textsc{SeS}}

\textsc{SeS} is a component evaluating the subjectivity of a sample while also taking into account the semantic relation between the answers. As such, the score it provides is a measure of \emph{reliability} of a sample (and of a dataset). Since a high reliable sample is one where annotators either converge on the same answer or pick up semantically related answers, we might take \textsc{SeS} as an indirect measure of dataset \emph{feasibility}: The higher the score assigned to a sample, the higher the probability to guess the correct answer. We test this intuition by analyzing VQA3+ accuracy against \textsc{SeS}. If \textsc{SeS} captures the degree of feasibility of a sample, we should observe a higher accuracy in correspondence to high values of our component. Our intuition is fully confirmed for VQA 1.0 (Figure~\ref{fig:diffbars}, left), where accuracies increase on par with \textsc{SeS}. In contrast, a different pattern is observed for VizWiz (right), where the highest accuracy is obtained in samples with moderate \textsc{SeS} and monotonically decreases with increasingly-reliable scores. This pattern, we conjecture, might be due to the low number of cases having high \textsc{SeS} in VizWiz.

\section{Discussion}

We proposed \textsc{MaSSeS}, a novel multi-component metric for the evaluation of VQA. We showed the potential of such evaluation tool for gaining a higher-level, fine-grained understanding of models and data. Crucially, our metric can be used one component at a time: \textsc{Ma} for evaluating model predictions only, \textsc{S} and \textsc{SeS} for analyzing the quantitative and semantic reliability of a dataset, respectively. Overall, \textsc{MaSSeS} provides a single accuracy score that makes it comparable to other metrics such as VQA3+ or WUPS. Further investigation is needed to explore the functioning of our metric with other VQA models, as well as the impact of using various word embeddings techniques and similarity thresholds on the overall score.

\section*{Acknowledgments}

A preliminary version of this work was presented at the ECCV2018 workshop on Shortcomings in Vision and Language (SiVL). In that venue, we had insightful discussions with Aishwarya Agrawal, Dhruv Batra, Danna Gurari, Stefan Lee, Vicente Ordonez, and many others. We thank them for helping us improving this manuscript.

{\small
\bibliographystyle{ieee}
\bibliography{egbib}
}

\end{document}